\documentclass[11pt,a4paper]{article}
\usepackage[hyperref]{acl2017}
\aclfinalcopy
\usepackage{times}
\usepackage{latexsym}
\usepackage{algorithm} 
\usepackage{algorithmic} 
\usepackage{multirow} 
\usepackage{amsmath}
\usepackage{xcolor}
\usepackage{epsfig}

\newcommand{\argmax}{\operatornamewithlimits{argmax}}

\title{Lock-Free Parallel Perceptron for Graph-based Dependency Parsing}

\author{Xu Sun \and Shuming Ma\\
MOE Key Laboratory of Computational Linguistics, Peking University\\
School of Electronics Engineering and Computer Science, Peking University\\
\{xusun, shumingma\}@pku.edu.cn\\
}

\begin{document}

\maketitle

\begin{abstract}
Dependency parsing is an important NLP task. A popular approach for dependency parsing is structured perceptron. Still, graph-based dependency parsing has the time complexity of $O(n^3)$, and it suffers from slow training. To deal with this problem, we propose a parallel algorithm called parallel perceptron. The parallel algorithm can make full use of a multi-core computer which saves a lot of training time. Based on experiments we observe that dependency parsing with parallel perceptron can achieve 8-fold faster training speed than traditional structured perceptron methods when using 10 threads, and with no loss at all in accuracy.
\end{abstract}

\section{Introduction}

Dependency parsing is an important task in natural language processing. It tries to match head-child pairs for the words in a sentence and forms a directed graph (a dependency tree). Former researchers have proposed various models to deal with this problem~\cite{Bohnet2010,McDonaldPereira2006}.

Structured perceptron is one of the most popular approaches for graph-based dependency parsing. It is first proposed by Collins~\shortcite{Collins2002} and McDonald et al.~\shortcite{McdonaldEA2005} first applied it to dependency parsing.
The model of McDonald is decoded with an efficient algorithm proposed by Eisner~\shortcite{Eisner1996} and they trained the model with structured perceptron as well as its variant Margin Infused Relaxed Algorithm (MIRA)~\cite{CrammerSinger2002,TaskarEA2004}. It proves that MIRA and structured perceptron are effective algorithms for graph-based dependency parsing. McDonald and Pereira~\shortcite{McDonaldPereira2006} extended it to a second-order model while Koo and Collins~\shortcite{KooCollins2010} developed a third-order model. They all used perceptron style methods to learn the parameters.

Recently, many models applied deep learning to dependency parsing. Titov and Henderson~\shortcite{TitovHenderson2007} first proposed a neural network model for transition-based dependency parsing. Chen and Manning~\shortcite{ChenManning2014} improved the performance of neural network dependency parsing algorithm while Le and Zuidema~\shortcite{LeZuidema2014} improved the parser with Inside-Outside Recursive Neural Network. However, those deep learning methods are very slow during training~\cite{Sun2016Asynchronous}.

To address those issues, we hope to implement a simple and very fast dependency parser, which can at the same time achieve state-of-the-art accuracies. To reach this target, we propose a lock-free parallel algorithm called lock-free parallel perceptron. We use lock-free parallel perceptron to train the parameters for dependency parsing. Although lots of studies implemented perceptron for dependency parsing, rare studies try to implement lock-free parallel algorithms. McDonald et al. \shortcite{Mcdonaldetal2010} proposed a distributed perceptron algorithm. Nevertheless, this parallel method is not a lock-free version on shared memory systems. To the best of our knowledge, our proposal is the first lock-free parallel version of perceptron learning.

Our contribution can be listed as follows:
\begin{itemize}
\item The proposed method can achieve 8-fold faster speed of training than the baseline system when using 10 threads, and without additional memory cost. 

\item We provide theoretical analysis of the parallel perceptron, and show that it is convergence even with the worst case of full delay. The theoretical analysis is for general lock-free parallel perceptron, not limited by this specific task of dependency parsing.
\end{itemize}

\section{Lock-Free Parallel Perceptron for Dependency Parsing}

The dataset can be denoted as $\{(x_i,y_i)\}_{i=1}^n$ while $x_i$ is input and $y_i$ is correct output. $GEN$ is a function which enumerates a set of candidates $GEN(x)$ for input $x$. $\Phi(x,y)$ is the feature vector corresponding to the input output pair $(x,y)$. Finally, the parameter vector is denoted as $\alpha$.

In structured perceptron, the score of an input output pair is calculated as follows:
\begin{equation}
s(x,y)=\Phi(x,y) \cdot \alpha
\end{equation}
The output of structured perceptron is to generate the structure $y'$ with the highest score in the candidate set $GEN(x)$.

In dependency parsing, the input $x$ is a sentence while the output $y$ is a dependency tree. An edge is denoted as $(i,j)$ with a head $i$ and its child $j$. Each edge has a feature representation denoted as $f(i,j)$ and the score of edge can be written as follows:
\begin{equation}
s(i,j) = \alpha \cdot f(i,j)
\end{equation}
Since the dependency tree is composed of edges, the score are as follows:
\begin{equation}
\small
s(x,y) = \sum_{(i,j) \in y}s(i,j) = \sum_{(i,j) \in y} \alpha \cdot f(i,j)
\end{equation}
\begin{equation}
\small
\Phi(x,y) = \sum_{(i,j) \in y}f(i,j)
\end{equation}

\begin{algorithm}[tb]
\small
   \caption{Lock-free parallel perceptron}\label{algo2}
\begin{algorithmic}[1]

 \STATE {\textbf{input}: Training examples $\{(x_i,y_i)\}_{i=1}^n$}
 \STATE {\textbf{initialize}: $\alpha = 0$}
 \REPEAT
 \FORALL {Parallelized threads}
 \STATE {Get a random example $(x_i,y_i)$}
 \STATE {$y' = argmax_{z \in GEN(x)}\Phi(x,y) \cdot \alpha$}
 \STATE {if $(y' \neq y)$ then $\alpha = \alpha + \Phi(x,y) - \Phi(x,y')$}
 \ENDFOR
 \UNTIL {Convergence}
 \STATE \RETURN {The averaged parameters $\alpha^*$}

\end{algorithmic}
\end{algorithm}

The proposed lock-free parallel perceptron is a variant of structured perceptron~\cite{SunIJCAI09,tkde/SunML13,SunArXiv2015}. We parallelize the decoding process of several examples and update the parameter vector on a shared memory system.
In each step, parallel perceptron finds out the dependency tree $y'$ with the highest score, and then updates the parameter vector immediately, without any lock of the shared memory. In typical parallel learning setting, the shared memory should be locked, so that no other threads can modify the model parameter when this thread is computing the update term. Hence, with the proposed method the learning can be fully parallelized. This is substantially different compared with the setting of McDonald et al.~\shortcite{Mcdonaldetal2010}, in which it is not lock-free parallel learning.

\section{Convergence Analysis of Lock-Free Parallel Perceptron}

For lock-free parallel learning, it is very important to analyze the convergence properties, because in most cases lock-free learning leads to divergence of the training (i.e., the training fails).
Here, we prove that lock-free parallel perceptron is convergent even with the worst case assumption. The challenge is that several threads may update and overwrite the parameter vector at the same time, so we have to prove the convergence.

We follow the definition in Collins's work \cite{Collins2002}. We write $\overline{GEN(x)}$ as all incorrect candidates generated by input $x$. We define that a training example is separable with margin $\delta > 0$ if $\exists U$ with $\lVert U \rVert = 1$ such that
\begin{equation}
\forall z \in \overline{GEN(x)}, U \cdot \Phi(x,y) - U \cdot \Phi(x,z) \geq \delta
\end{equation}

Since multiple threads are running at the same time in lock-free parallel perceptron training, the convergence speed is highly related to the delay of update. Lock-free learning has update delay, so that the update term may be applied on a ``old'' parameter vector, because this vector may have already be modified by other threads (because it is lock-free) and the current thread does not know that. Our analysis show that the perceptron learning is still convergent, even with the worst case that all of the $k$ threads are delayed. To our knowledge, this is the first convergence analysis for lock-free parallel learning of perceptrons.


We first analyze the convergence of the worse case (full delay of update). Then, we analyze the convergence of optimal case (minimal delay). In experiments we will show that the real-world application is close to the optimal case of minimal delay.

\subsection{Worst Case Convergence}

Suppose we have $k$ threads and we use $j$ to denote the $j$'th thread, each thread updates the parameter vector as follows:
\begin{equation}
y_j' = \argmax_{z \in GEN(x)} {\Phi_j(x,y) \cdot \alpha}
\end{equation}
Recall that the update is as follows:
\begin{equation}
\alpha^{i+1} = \alpha^{i} + \Phi_j(x,y) - \Phi_j(x,y_j')
\end{equation}
Here, $y_j'$ and $\Phi_j(x,y)$ are both corresponding to $j^{th}$ thread while $\alpha^{i}$ is the parameter vector after $i^{th}$ time stamp.

Since we adopt lock-free parallel setting, we suppose there are $k$ perceptron updates in parallel in each time stamp. Then, after a time step, the overall parameters are updated as follows:
\begin{equation}
\alpha^{t+1} = \alpha^t + \sum_{j=1}^{k}(\Phi_j(x,y) - \Phi_j(x,y_j'))
\end{equation}
Hence, it goes to:
\begin{equation*}
\begin{split}
& U \cdot \alpha^{t+1} = U \cdot \alpha^t + \sum_{j=1}^{k}U \cdot (\Phi_j(x,y) - \Phi_j(x,y_j'))\\
&\geq U \cdot \alpha^t + k \delta
\end{split}
\end{equation*}
where $\delta$ is the separable margin of data, following the same definition of Collins \shortcite{Collins2002}.
Since the initial parameter $\alpha = 0$, we will have that $U \cdot \alpha^{t+1} \geq tk\delta$ after $t$ time steps. Because $U \cdot \alpha^{t+1} \leq \lVert U \rVert \lVert \alpha^{t+1} \rVert$, we can see that
\begin{equation}\label{eq102}
\lVert \alpha^{t+1} \rVert \geq tk\delta
\end{equation}

On the other hand, $\lVert \alpha^{t+1} \rVert$ can be written as:
\begin{equation*}
\begin{split}
\lVert \alpha^{t+1} \rVert^2 & =  \lVert \alpha^{t} \rVert^2 + \lVert \sum_{j=1}^{k}(\Phi_j(x,y) - \Phi_j(x,y_j')) \rVert^2 \\
& +2\alpha^{t} \cdot (\sum_{j=1}^{k}(\Phi_j(x,y) - \Phi_j(x,y_j')))  \\
& \leq \lVert \alpha^{t} \rVert^2 + k^2 R^2
\end{split}
\end{equation*}
where $R$ is the same definition following Collins \shortcite{Collins2002} such that $\Phi(x,y) - \Phi(x,y_j') \leq R$.
The last inequality is based on the property of perceptron update such that the incorrect score is always higher than the correct score (the searched incorrect structure has the highest score) when an update happens. Thus, it goes to:
\begin{equation}\label{eq101}
\lVert \alpha^{t+1} \rVert^2 \leq t k^2 R^2
\end{equation}

Combining Eq.\ref{eq101} and Eq.\ref{eq102}, we have:
\begin{equation}
t^2 k^2 \delta^2 \leq \lVert \alpha^{t+1} \rVert^2 \leq tk^2R^2
\end{equation}
Hence, we have:
\begin{equation}
t \leq R^2/\delta^2
\end{equation}

This proves that the lock-free parallel perceptron has bounded number of time steps before convergence even with the worst case of full delay, and the number of time steps is bounded by $t \leq R^2/\delta^2$ in the worst case. The worst case means that the parallel perceptron is convergent even if the update is extremely delayed, such that $k$ threads are updating based on the same old parameter vector.

\subsection{Optimal Case Convergence}
In practice the worst case of extremely delayed update is not probable to happen, or at least not always happening. Thus, we expect that the real convergence speed should be much faster than this worst case bound. The optimal bound is as follows:
\begin{equation}
t \leq R^2/(k\delta^2)
\end{equation}
This bound is derived when the parallel update is not delayed (i.e., the update of each thread is based on a most recent parameter vector). As we can see, in the optimal case we can get $k$ times speed up by using $k$ threads for lock-free parallel perceptron training. This can achieve full acceleration of training by using parallel learning.

\begin{table}[tb]
\begin{tabular}{|c|c|c|}
\hline
Models & 1st-order & 2nd-order \\
\hline
MST Parser & 91.60 & 92.30 \\


Locked Para-Perc & 91.68 & \textbf{92.55} \\
\hline \hline
Lock-free Para-Perc 5-thread & 91.70 & \textbf{92.55} \\
\hline
Lock-free Para-Perc  10-thread & \textbf{91.72} & 92.53 \\
\hline
\end{tabular}
\caption{Accuracy of baselines and our method.} \label{table1}
\end{table}

\begin{table}[tb]
\begin{tabular}{l|r|r}
\hline
Models & 1st-order & 2nd-order \\
\hline
Structured Perc & 1.0x(449s) & 1.0x(3044s) \\

Locked Para-Perc & 5.1x(88s) & 5.0x(609s) \\

\hline \hline
Lock-free Para-Perc 5-thr. & 4.3x(105s) & 4.5x(672s) \\
\hline
Lock-free Para-Perc 10-thr. & \textbf{8.1x(55.4s)} & \textbf{8.3x(367s)} \\
\hline
\end{tabular}
\caption{Speed up and time cost per pass of our algorithm}\label{table2}
\end{table}

\begin{figure*}[tb]
\centering
\begin{tabular}{cccc}

\epsfig{file=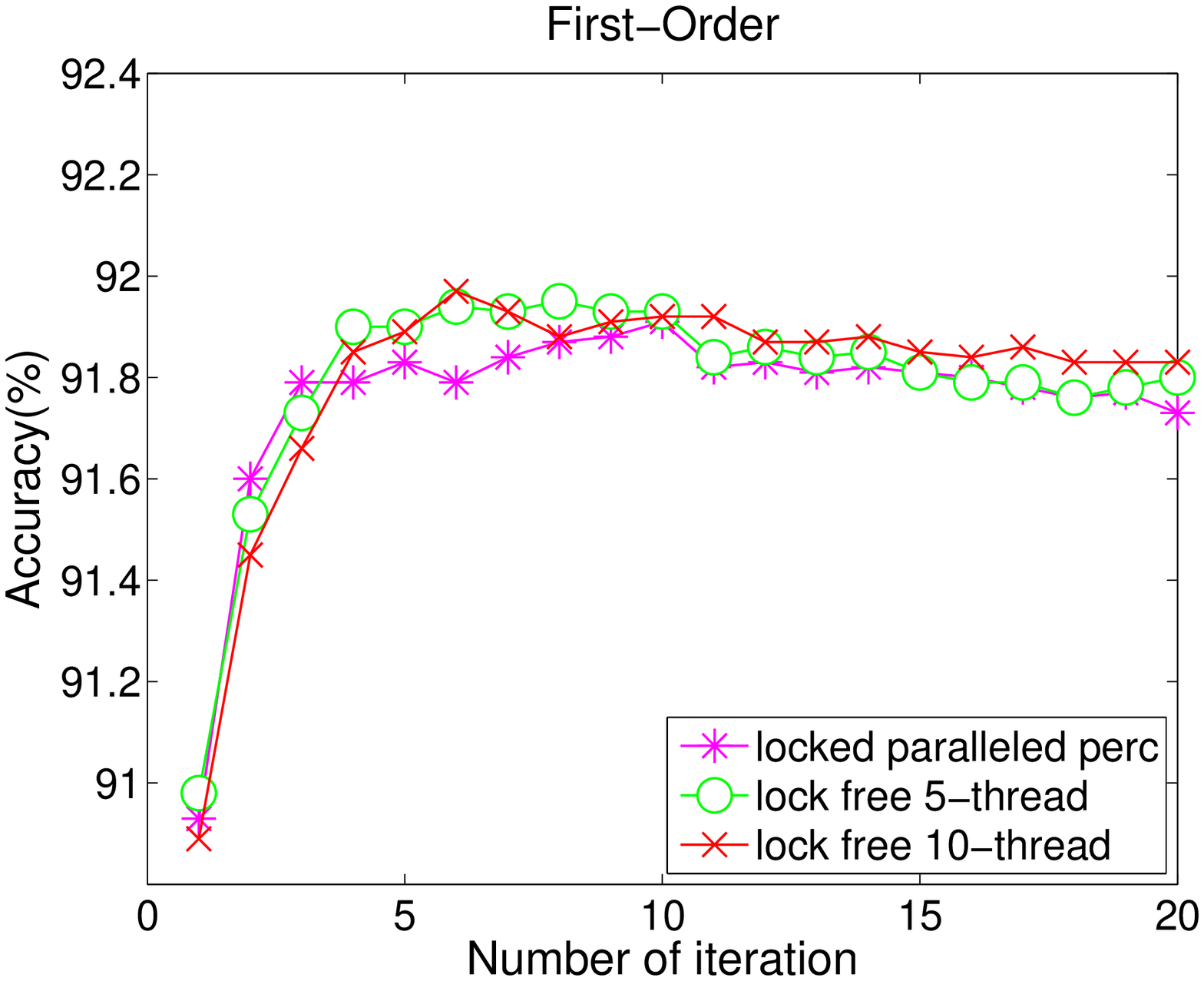,width=0.4\linewidth,clip=} &
\epsfig{file=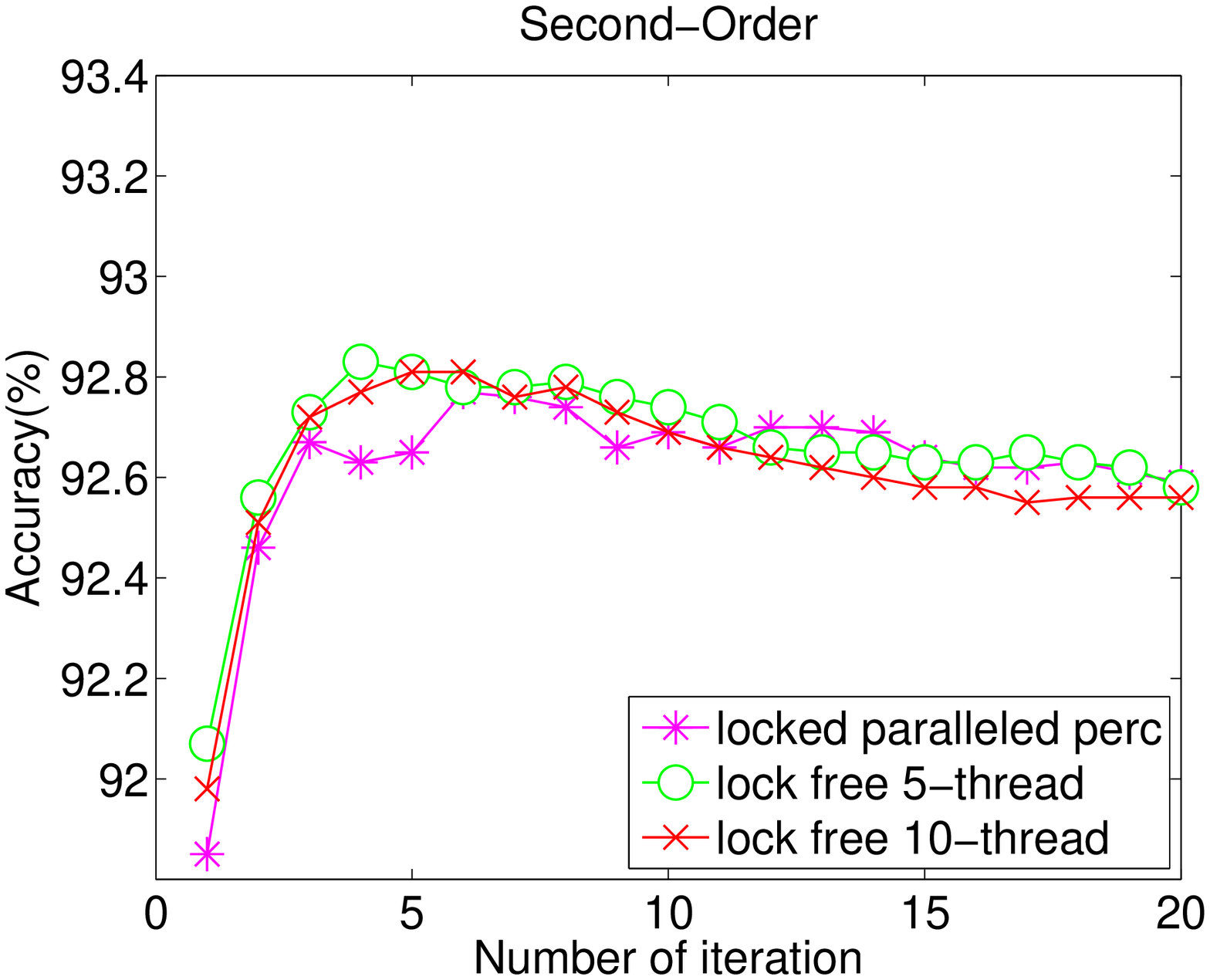,width=0.4\linewidth,clip=}\\

\end{tabular}
\caption{Accuracy of different methods for dependency parsing.
}\label{fig1}
\vspace{-0.1in}
\end{figure*}

\begin{figure*}[tb]
\centering
\begin{tabular}{cccc}
\epsfig{file=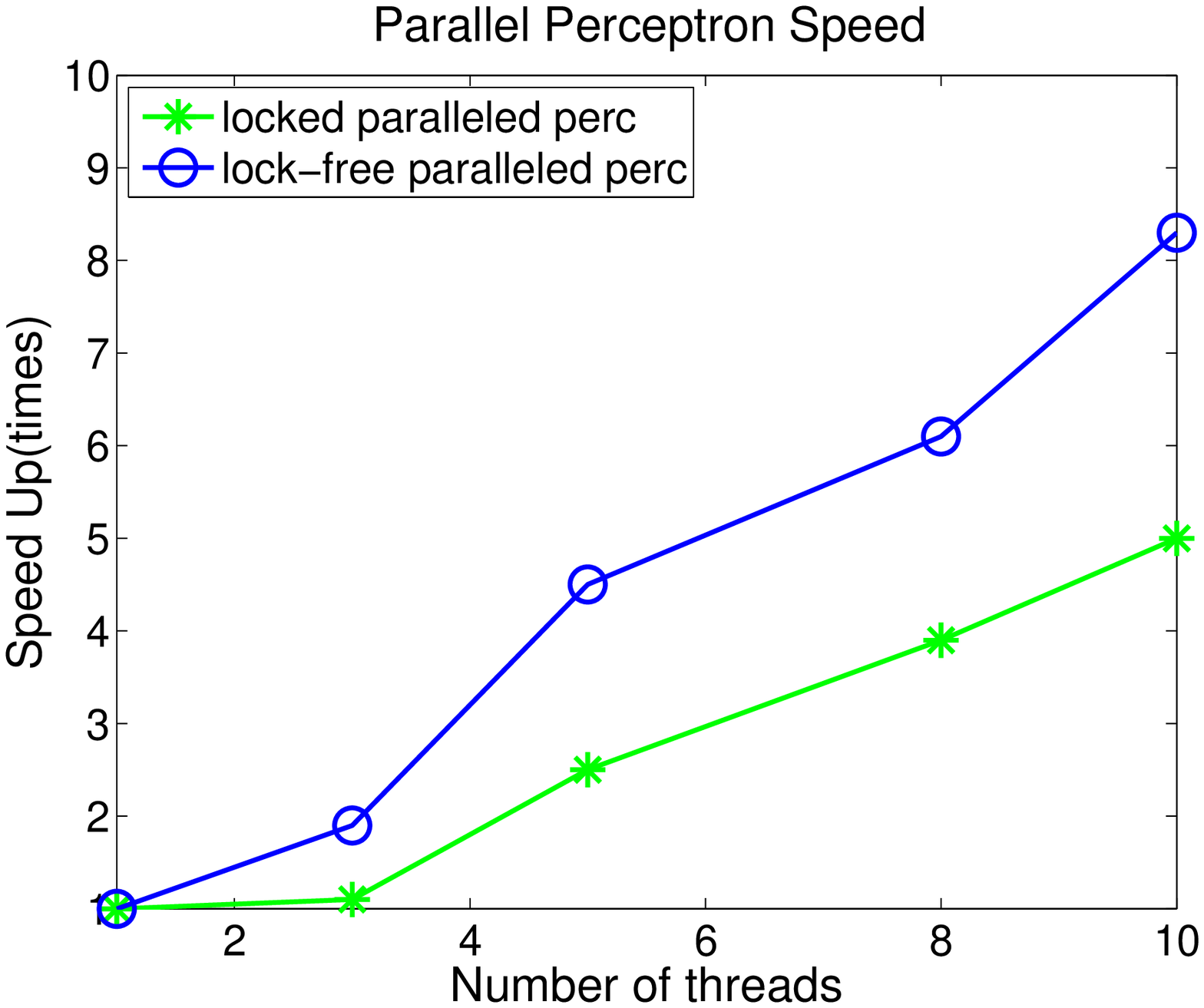,width=0.4\linewidth,clip=}  &
\epsfig{file=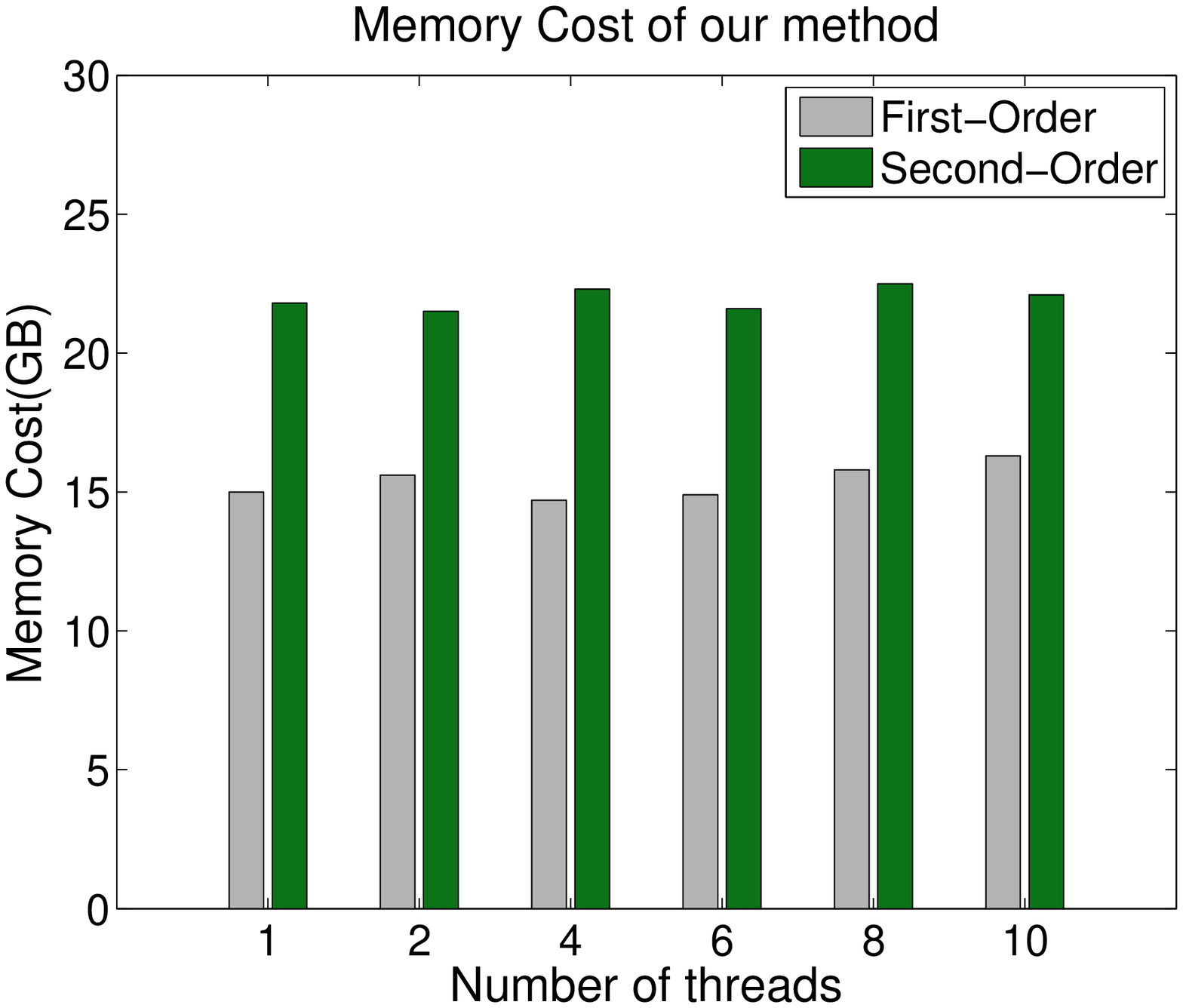,width=0.4\linewidth,clip=} \\

\end{tabular}
\caption{Speed up and memory cost of different methods for dependency parsing.
}\label{fig2}
\vspace{-0.1in}
\end{figure*}

\section{Experiments}

\subsection{Dataset}

Following prior work, we use English Penn TreeBank (PTB)~\cite{Marcusetal1993} to evaluate our proposed approach. We follow the standard split of the corpus, using section 2-21 as training set, section 22 as development set, and section 23 as final test set. 
We implement two popular model of graph-based dependency parsing: first-order model and second-order model. We tune all of the hyper parameters in development set. The features in our model can be found in McDonald et al.~\shortcite{McdonaldEA2005,McDonaldPereira2006}.
Our baselines are traditional perceptron, MST-Parser~\cite{McdonaldEA2005}\footnote{www.seas.upenn.edu/\~strctlrn/MSTParser/MSTParser.html}, 
and the locked version of parallel perceptron. All of the experiment is conducted on a computer with the Intel(R) Xeon(R) 3.0GHz CPU.

\subsection{Results}

Table~\ref{table2} shows that our lock-free method can achieve 8-fold faster speed than the baseline system, which is better speed up when compared with locked parallel perceptron. 
For both 1st-order parsing and 2nd-order parsing, the results are consistent that the proposed lock-free method achieves the best rate of speed up. The results show that the lock-free parallel peceptron in real-world applications is near the optimal case theoretical analysis of low delay, rather than the worst case theoretical analysis of high delay.

The experimental results of accuracy are shown in Table~\ref{table1}. The baseline MST-Parser~\cite{McdonaldEA2005} is a popular system for dependency parsing. Table~\ref{table1} shows that our method with 10 threads outperforms the system with single-thread. Our lock system is slightly better than MST-Parser mainly because we use more feature including distance based feature -- our distance features are based on larger size of contextual window. 

Figure~\ref{fig1} shows that the lock-free parallel perceptron has no loss at all on parsing accuracy on both 1st-order and 2nd-order parsing setting, in spite of the substantial speed up of training. 

Figure~\ref{fig2} shows that the method can achieve near linear speed up, and with almost no extra memory cost. 

\section{Conclusions}

We propose lock-free parallel perceptron for graph-based dependency parsing. Our experiment shows that it can achieve more than 8-fold faster speed than the baseline when using 10 running threads, and with no loss in accuracy. We also provided convergence analysis for lock-free parallel perceptron, and show that it is convergent in the lock-free learning setting. The lock-free parallel perceptron can be directly used for other structured prediction NLP tasks.

\section{Acknowledgements}

This work was supported in part by National Natural Science Foundation of China (No. 61673028), and National High Technology Research and Development Program of China (863 Program, No. 2015AA015404). Xu Sun is the corresponding author of this paper.

\bibliographystyle{acl_natbib}
\bibliography{bib}

\end{document}